\documentclass[runningheads]{llncs}

\usepackage[nogeometry,noformatting]{stdcommands} %% Humberto's awesome command package
\usepackage{hyperref} %% enable active PDF urls and \hypersetup
\usepackage{graphicx} %% enable \includegraphics
\usepackage[utf8]{inputenc} %% setup source input encoding
\usepackage{color} %% enable \color
\usepackage{algorithmicx} %% enable algorithmic environment
\usepackage[noend]{algpseudocode} %% enable standard algortihmicx layout
\usepackage{xspace} %% enable smart do-we-need-a-space? hack \xspace
\usepackage{multirow} %% enable \multirow in tabular environments
\usepackage{array} %% enable using "m" and "b" specifications in tabular environment
\usepackage[indention=.5em]{subcaption} %% enable subfigure environment

\hypersetup{%
  pdfinfo={
    Title={MCA-based Rule Mining Enables Interpretable Inference in Clinical Psychiatry},
    Author={Qingzhu Gao, Humberto Gonzalez, Parvez Ahammad}
  }
}

\newcommand{\review}[1]{#1}
\newcommand{\true}{\emph{True}\xspace}
\newcommand{\false}{\emph{False}\xspace}
\newcommand{\cnpq}[2]{\emph{#1}\##2}
\DeclareMathOperator{\supp}{supp}
\algnewcommand{\AlgoAnd}{\textbf{and}\xspace}
\algnewcommand{\AlgoOr}{\textbf{or}\xspace}
\algnewcommand{\AlgoContinue}{\textbf{continue}\xspace}

\begin{document}

%%%%%
%% Title and Author information
%%%%%
\title{%
  MCA-based Rule Mining Enables Interpretable Inference in Clinical Psychiatry%
}
\titlerunning{%
  MCA-based Rule Mining Enables Interpretable Inference in Clinical Psych.%
}
\author{%
  Qingzhu Gao\inst{1,2} \and%\orcidID{0000-0002-9510-7411}
  Humberto Gonzalez\inst{1,2} \and%\orcidID{0000-0002-9883-2053}
  Parvez Ahammad\inst{1}%\orcidID{0000-0003-1536-1207}
}
\institute{%
  BlackThorn Therapeutics, San Francisco, CA 94103 \and%
  Equal contribution authors%
}

\maketitle

%%%%%
%% Contents
%%%%%
\begin{abstract}
  Development of interpretable machine learning models for clinical healthcare applications has the potential of changing the way we understand, treat, and ultimately cure, diseases and disorders in many areas of medicine.
  These models can serve not only as sources of predictions and estimates, but also as discovery tools for clinicians and researchers to reveal new knowledge from the data.
  High dimensionality of patient information (e.g., phenotype, genotype, and medical history), lack of objective measurements, and the heterogeneity in patient populations often create significant challenges in developing interpretable machine learning models for clinical psychiatry in practice.

  In this paper we take a step towards the development of such interpretable models.
  First, by developing a novel categorical rule mining method based on Multivariate Correspondence Analysis~(MCA) capable of handling datasets with large numbers of features, and second, by applying this method to build transdiagnostic Bayesian Rule List models to screen for psychiatric disorders using the Consortium for Neuropsychiatric Phenomics dataset.
  We show that our method is not only at least 100~times faster than state-of-the-art rule mining techniques for datasets with 50~features, but also provides interpretability and comparable prediction accuracy across several benchmark datasets.
\end{abstract}

\section{Introduction}
\label{sec:introduction}

The use of novel Artificial Intelligence (AI) tools to derive insights from clinical psychiatry datasets has consistently increased in recent years~\cite{beam2018big}, generating highly predictive models for heterogeneous datasets. %, mostly due to the prevalence of algorithms capable of analyzing heterogeneous datasets and at the same time producing highly predictive models.
While high predictability is indeed a desirable result, the healthcare community requires that the AI models are also \emph{interpretable}, so that experts can learn new insights from these models, or even better, so that experts can improve the performance of the models by tuning the data-driven models.
We take a practical approach towards solving this problem, by developing a new rule mining method for wide categorical datasets, and \review{by applying our mining method to build interpretable transdiagnostic screening tools for psychiatric disorders - aiming to capture underlying commonalities among these disorders.}
% In this paper we take a practical approach, both developing a new algorithm capable of mining association rules from wide categorical datasets, and applying our mining method towards \review{building models that are both predictive and interpretable, resulting in transdiagnostic screening tools for psychiatric disorders, i.e., screening tools that capture underlying commonalities among psychiatric disorders}.

% The artificial intelligence research community has been urged to develop interpretable machine learning methods, which can provide accessible and explicit explanations.
\review{Starting from early clinical decision support systems (CDSS)~\cite{wyatt1991field}, the interpretation that clinicians obtain from data-driven models was identified as a critical element in their practical deployment}.
A report by the AI~Now Institute remarks as the top recommendation in their 2017 report that core government agencies, including those responsible for healthcare, ``should no longer use \emph{black box} AI and algorithmic systems''~\cite{campolo2017ai}.
The Explainable Artificial Intelligence (XAI) program at DARPA has as one of its goals to ``enable human users to understand, appropriately trust, and effectively manage the emerging generation of artificially intelligent partners''~\cite{gunning2017explainable}.
In contrast, popular machine learning methods such as artificial neural networks~\cite{lecun2015deep} and ensemble models~\cite{dietterich2000experimental} are known for their elusive readout.
For example, while artificial neural network applications exist for tumor detection in CT scans~\cite{anthimopoulos2016lung}, it is virtually impossible for a person to understand the rational behind such a mathematical abstraction.
% As machine learning researchers, we aim to develop an algorithms that can explain itself, while providing excellence predictions.

Interpretability is often loosely defined as understanding not only \emph{what} a model emitted, but also \emph{why} it did~\cite{gilpin2018explaining}.
% In this context, linguistic explanations are considered as some of the most interpretable when compared to examining the internal representations of complex models. %~\cite{morcos2018on}.
\review{As explained in~\cite{lipton2018mythos}, rule-based decision models offer desirable interpretation properties such as \emph{trust}, \emph{transparent simulatability}, and \emph{post-hoc text explanations}.}
% A simple well-pruned decision tree~\cite{quinlan1986induction} can be expressed in natural language, however, its simplicity limits the predictive power.
Recent efforts towards interpretable machine learning models in healthcare can be found in the literature, such as the development of a boosting method to create decision trees as the combination of single decision nodes~\cite{valdes2016mediboost}.
Bayesian Rule List~(BRL) \cite{rudin2013learning,letham2015interpretable} mixes the interpretability of sequenced logical rules for categorical datasets, together with the inference power of Bayesian statistics.
Compared to decision trees, BRL rule lists take the form of a hierarchical series of \emph{if-then-else} statements where model emissions are correspond to the successful association to a given rule.
BRL results in models that are inspired, and therefore similar, to standard human-built decision-making algorithms.

While BRL is by itself an interesting model to try on clinical psychiatry datasets, it relies on the existence of an initial set of rules from which the actual rule lists are built, which is similar to the approach taken by other associative classification methods~\cite{liu1998integrating,yin2003cpar,li2001cmar}.
Frequent pattern mining has been a standard tool to build such initial set of rules, with methods like Apriori~\cite{agrawal1994fast} and FP-Growth~\cite{han2000mining} being commonly used to extract rules from categorical datasets.
However, frequent pattern mining methods do not scale well for wide datasets, i.e., datasets where the total number of categorical features is much larger than the number of samples, commonly denoted as $p \gg n$.
Most clinical healthcare datasets are wide and thus require new mining methods to enable the use of BRL in this research area.

In this paper we propose a new rule mining technique that is not based on the frequency in which certain categories simultaneously appear.
Instead, we use Multiple Correspondence Analysis (MCA)~\cite{greenacre2006multiple}, a particular application of correspondence analysis to categorical datasets, to establish a similarity score between different associative rules.
We show that our new MCA-miner method is significantly faster than commonly used frequent pattern mining methods, and that it scales well to wide datasets.
Moreover, we show that MCA-miner performs equally well as other miners when used together with BRL.
Finally, we use MCA-miner and BRL to analyze \review{a dataset designed for the transdiagnostic study of psychiatric disorders, building interpretable predictors to support clinician screening tasks}.

% The remainder of this paper is organized as follows.
% In Sec.~\ref{sec:problem_description}, we layout the problem of constructing a rule based classifier. % and establish the notation we use throughout the paper.
% We illustrate the algorithmic structure of MCA-miner in Sec.~\ref{sec:method}. %, and discuss the mathematical property of pruning rules based on coordinates scoring.
% We compare the performance of our new method against standard benchmark datasets in Sec.~\ref{sec:uci}, and we study new screening data-driven models for psychiatric disorders in Sec.~\ref{sec:cnp}.

% %% Associative Rule Learning lit review
% FP-Growth~\cite{han2000mining}.

% Rule based MDD monitoring~\cite{lin2018databased}
\section{Problem Description and Definitions}
\label{sec:problem_description}

We begin by introducing definitions used throughout this paper.
An \emph{attribute}, denoted $a$, is a categorical property of each data sample, which can take a discrete and finite number of values denoted as $\abs{a}$.
A \emph{literal} is a Boolean statement checking if an attribute takes a given value, e.g., given an attribute $a$ with categorical values $\set{c_1, c_2}$ we can define the following literals: \emph{$a$ is $c_1$}, and \emph{$a$ is $c_2$}.
Given a collection of attributes $\set{a_i}_{i=1}^p$, a \emph{data sample} is a list of categorical values, one per attribute.
A \emph{rule}, denoted $r$, is a collection of literals, with length $\abs{r}$, which is used to produce Boolean evaluations of data samples as follows: a rule evaluates to \true whenever all the literals are also \true, and evaluates to \false otherwise.

In this paper we consider the problem of efficiently building \emph{rule lists}, which are evaluated sequentially until one rule is satisfied, for datasets with a large total number of categories among all attributes (i.e., $\sum_{i=1}^p \abs{a_i}$), a common situation among datasets related to health care or pharmacology.
Given $n$ data samples, we represent a dataset as a matrix $X$ with dimensions $n \times p$, where $X_{i,j}$ is the category assigned to the $i$-th sample for the $j$-th attribute.
We also consider a categorical label for each data sample, collectively represented as a vector $Y$ with length $n$.
We denote the number of label categories by $\ell$, where $\ell \geq 2$.
If $\ell = 2$, we are solving a standard binary classification problem.
If, instead, $\ell > 2$ then we solve a multi-class classification problem. %, as shown in Sec.~\ref{sec:cnp}.

% \subsection{Bayesian Rule Lists}
% \label{sec:brl}

Bayesian Rule Lists (BRL) is a framework proposed by Rudin et al.~\cite{rudin2013learning,letham2015interpretable} to build interpretable classifiers.
% An example of a BRL output trained on the commonly used Titanic survival dataset~\cite{hendricks2015titanic}, as shown in~\cite{letham2015interpretable}, is included in Fig.~\ref{fig:titanic}.
% \begin{figure}[tp]
%   \centering
%   \begin{algorithmic}[1]
%     \If{sex is male \AlgoAnd age is adult} $\theta = 0.21$
%     \ElsIf{class is 3rd} $\theta = 0.44$
%     \ElsIf{class is 1st} $\theta = 0.96$
%     \Else{} $\theta = 0.88$
%     \EndIf
%   \end{algorithmic}
%   \caption{%
%     BRL output on the Titanic survival dataset as reported in~\cite{letham2015interpretable}.
%     $\theta$ denotes the probability of survival.
%   }
%   \label{fig:titanic}
% \end{figure}
Although BRL is a significant step forward in the development of XAI methods, searching over the configuration space of all possible rules containing all possible combinations of literals obtained from a given dataset is simply infeasible.
Letham et al.~\cite{letham2015interpretable} offer a good compromise solution to this problem, where first a set of rules is mined from a dataset, and then BRL searches over the configuration space of combinations of the prescribed set of rules using a custom-built MCMC algorithm.
While efficient rule mining methods are available in the literature, we show in Sec.~\ref{sec:cnp} that such methods fail to execute on datasets with a large total number of categories, due to either unacceptably long computation time or prohibitively high memory usage.

In this paper we build upon the method in~\cite{letham2015interpretable} developing two improvements.
First, we propose a novel rule mining algorithm based on Multiple Correspondence Analysis that is both computational and memory efficient, enabling us to apply BRL on datasets with a large total number of categories.
Our MCA-based rule mining algorithm is explained in detail in Sec.~\ref{sec:method}.
Second, we parallelized the MCMC search method in BRL by executing individual Markov chains in separate CPU cores of a computer.
Moreover, we periodically check the convergence of the multiple chains using the generalized Gelman \& Rubin convergence criteria~\cite{brooks1998general,gelman1992inference}, thus stopping the execution once the convergence criteria is met.
As shown \review{in Sec.~\ref{sec:cnp_perf}}, our implementation is significantly faster than the original single-core version, enabling the study of more datasets with longer rules or a large number of features.

\section{MCA-based Rule Mining}
\label{sec:method}

Multiple Correspondence Analysis (MCA)~\cite{greenacre2006multiple} is a method that applies the power of Correspondence Analysis (CA) to categorical datasets.
For the purpose of this paper it is important to note that MCA is the application of CA to the indicator matrix of all categories in the set of attributes, thus generating principal vectors projecting each of those categories into a euclidean space.
We use these principal vectors to build an efficient heuristic merit function over the set of all available rules given the categories in a dataset.
% Moreover, the structure of our merit function enables us to efficiently mine the best rules, as detailed below.

\subsection{Rule Score Calculation}
First, we compute the MCA principal vectors of the extended data matrix concatenating $X$ and $Y$, defined as $Z = \smat{X & Y}$ with dimensions $n \times (p+1)$.
% We then compute the MCA principal vectors for each category present of $Z$.
Let us denote the MCA principal vectors associated each categorical value by $\set{v_j}_{j=1}^{\sum_i \abs{a_i}}$, where $\set{a_i}_{i=1}^p$ is the set of attributes in the dataset $X$.
Also, let us denote the MCA principal vectors associated to label categories by $\set{\omega_k}_{k=1}^\ell$.

Since each category can be mapped to a literal statement, as explained in Sec.~\ref{sec:problem_description}, these principal vectors serve as a heuristic to evaluate the quality of a given literal to predict a label~\cite{zhu2010feature}.
Therefore, we define the score between each $v_j$ and each $\omega_k$ by $\rho_{j,k} = \cos \angle\p{v_j,\omega_k} = \frac{\dprod{v_j}{\omega_k}}{\norm{v_j}_2\, \norm{\omega_k}_2}$.
% \begin{equation}
%   \label{eq:angle_score}
%   \rho_{j,k} = \cos \angle\p{v_j,\omega_k} = \frac{\dprod{v_j}{\omega_k}}{\norm{v_j}_2\, \norm{\omega_k}_2}.
% \end{equation}
Note that in the context of random variables, $\rho_{i,k}$ is equivalent to the correlation between $v_j$ and $\omega_k$~\cite{loeve1977probability}.

We compute the \emph{score between a rule $r$ and label category $k$}, denoted $\mu_k(r)$, as the average among the scores between the literals in $r$ and the same label category: $\mu_k(r) = \frac{1}{\abs{r}}\, \sum_{l \in r} \rho_{l,k}$.
% \begin{equation}
%   \label{eq:rule_score}
%   \mu_k(r) = \frac{1}{\abs{r}}\, \sum_{l \in r} \rho_{l,k}.
% \end{equation}
Finally, we search the configuration space of rules $r$ built using the combinations of all available literals in a dataset such that $\abs{r} \leq r_{\text{max}}$, and identify those with highest scores for each label category.
These top rules are the output of our miner, and are passed over to the BRL method as the set of rules from which rule lists will be built.

The pseudocode for our rule mining algorithm is shown in Fig.~\ref{fig:mca_miner}, where we parallelized the loop iterating over label categories in line~\ref{algo:mca_for_label}. % can be easily parallelized as a multi-core computation, significantly reducing the mining time \review{as shown in Sec.~\ref{sec:cnp_perf}}.

\begin{figure}[t]
  \centering
  \begin{algorithmic}[1]
    \Require Dataset $X$ with labels $Y$.
    User parameters $r_{\text{max}}$, $s_{\text{min}}$, $\mu_{\text{min}}$, and $M$.
    \State Compute literal scores $\rho_{l,k}$ for each literal $l$ in $X$ and label category $k$ in $Y$
    \State $R = \emptyset$
    \For{label $k \in Y$} \label{algo:mca_for_label}
    \State $R_k = \emptyset$
    \For{literal $l \in X$}
    \If{$\rho_{l,k} \geq \mu_{\text{min}}$ \AlgoAnd $\supp_k(l) \geq s_{\text{min}}$}
    \State $R_k \gets R_k \cup \set{l}$
    \EndIf
    \EndFor
    \For{rule length $\eta \in \set{1,\ldots,r_{\text{max}}-1}$}
    \For{rule $r \in R_k$ with $\abs{r} = \eta$}
    \State $\mu_{\text{min}} \gets M\text{-th top score in}\ R_k$
    \If{$\mu_k\p{r} < m_k(\eta)$}
    \AlgoContinue
    \EndIf
    \For{literal $l \in X$}
    \State $\hat{r} \gets r \cup l$
    \If{$\mu_k\p{\hat{r}} \geq \mu_{\text{min}}$ \AlgoAnd $\supp_k\p{\hat{r}} \geq s_{\text{min}}$}
    \State $R_k \gets R_k \cup \set{\hat{r}}$
    \EndIf
    \EndFor
    \EndFor
    \EndFor
    \State $R \gets R \cup \set{\text{top $M$ rules in $R_k$ sorted by score}}$
    \EndFor
    \State \Return $R$
  \end{algorithmic}
  \caption{Pseudocode of our MCA-based rule mining algorithm.}
  \label{fig:mca_miner}
\end{figure}

\subsection{Rule Prunning}
Since the number of rules generated by all combinations of all available literals up to length $r_{\text{max}}$ is excessively large even for modest values of $r_{\text{max}}$, our miner includes two conditions under which we efficiently eliminate rules from consideration.

First, similar to the approach in FP-Growth~\cite{han2000mining} and other popular miners, we eliminate rules whose support over each label category is smaller than a user-defined threshold $s_{\text{min}}$.
Recall that the \emph{support} of a rule $r$ for label category $k$, denoted $\supp_k(r)$, is the fraction of data samples that the rule evaluates to \true among the total number of data samples associated to a given label.
% Given a rule $r$, note that the support of every other rule $\hat{r}$ containing the collection of literals in $r$ satisfies $\supp_k\p{\hat{r}} \leq \supp_k(r)$.
% Hence, once a rule $r$ fails to pass our minimum support test, we stop considering all rules longer than $r$ that also contain the all the literals in $r$.
Given a rule $r$, note that once a rule $r$ fails to pass our minimum support test, we stop considering all rules longer than $r$ that also contain the all the literals in $r$ since their support is necessarily smaller.

Second, we eliminate rules whose score is smaller than a user-defined threshold $\mu_{\text{min}}$.
Suppose that we want to build a new rule $\hat{r}$ by taking a rule $r$ and adding a literal $l$.
In that case, given a category $k$ the score of this rule must satisfy $\mu_k\p{\hat{r}} = \frac{1}{\abs{r}+1}\, \pb{\abs{r}\, \mu_k(r) + \rho_{l,k}} \geq \mu_{\text{min}}$.
% \begin{equation}
%   \mu_k\p{\hat{r}} = \frac{\abs{r}\, \mu_k(r) + \rho_{l,k}}{\abs{r}+1} \geq \mu_{\text{min}}.
% \end{equation}
Let $\overline{\rho}_k = \max_l \rho_{l,k}$ be the largest score among all available literals, then we can predict that at least one extension of $r$ will have a score greater than $\mu_{\text{min}}$ if $\mu_k(r) \geq \frac{1}{\abs{r}}\, \pb{\p{\abs{r} + 1}\, \mu_{\text{min}} - \overline{\rho}_k} = m_k(\abs{r})$.
% \begin{equation}
%   \label{eq:rule_score_min}
%   \mu_k(r) \geq \frac{\p{\abs{r} + 1}\, \mu_{\text{min}} - \overline{\rho}_k}{\abs{r}} = m_k(\abs{r}).
% \end{equation}
Given the maximum number of rules to be mined per label $M$, we recompute $\mu_{\text{min}}$ as we iterate combining literals to build new rules.
% Indeed, we periodically sort the scores for our temporary list of candidate rules and set $\mu_{\text{min}}$ equal to the score of the $M$-th rule in the sorted list.
As $\mu_{\text{min}}$ increases due to better candidate rules becoming available, the rule-acceptance bound $m_k$ becomes more restrictive, resulting in less rules being considered and therefore in a faster overall mining.

\section{Benchmark Experiments}
\label{sec:uci}

% Our MCA-miner method, when used together with BRL, offers the power of rule list interpretability while maintaining the predictive capabilities of already established machine learning methods.

We benchmark the performance and computational efficiency of our MCA-miner against the ``Titanic'' dataset~\cite{hendricks2015titanic}, as well as the following 5~datasets available in the UCI Machine Learning Repository~\cite{dua2017uci}: ``Adult,'' ``Autism Screening Adult'' (\emph{ASD}), ``Breast Cancer Wisconsin (Diagnostic)'' (\emph{Cancer}),  ``Heart Disease'' (\emph{Heart}), and ``HIV-1 protease cleavage'' (\emph{HIV}). These datasets represent a wide variety of real-world experiments and observations, thus allowing us to fairly compare our improvements against the original BRL implementation using the FP-Growth miner.
All 6~benchmark datasets correspond to binary classification tasks.
We conduct the experiments using the same set up in each of the benchmarks, namely quantizing all continuous attributes into either 2~or 3~categories, while keeping the original categories of all other variables.
% First, we transform the dataset into a format that is compatible with our BRL implementation.
% Second, we quantize all continuous attributes into either 2~or 3~categories, while keeping the original categories of all other variables.
% It is worth noting that depending on the dataset and how its data was originally collected, we prioritize the existing taxonomy and expert domain knowledge to generate the continuous variable quantization.
% We simply generate a balanced quantization when no other information was available.
We train and test each model using 5-fold cross-validations, reporting the average \emph{accuracy} and \emph{Area Under the ROC Curve} (ROC-AUC) as model performance measurements.

\begin{table}[t]
  \centering
  \caption{%
    Performance evaluation of FP-Growth against MCA-miner when used with BRL on benchmark datasets.
    $t_{\text{train}}$ is the full training wall time in seconds.
  }
  \label{tab:uci_table}
  \begin{tabular}{c||c|c|c|c|c|c|c|c|c}
    \multirow{2}{*}{Dataset} & \multirow{2}{*}{$n$}
    & \multirow{2}{*}{$p$} & \multirow{2}{*}{$\sum_{i=1}^p \abs{a_i}$}
    & \multicolumn{3}{c|}{FP-Growth + BRL} & \multicolumn{3}{c}{MCA-miner + BRL}\\
    \cline{5-10}
    & & & & Accuracy & ROC-AUC & $t_{\text{train}}$ & Accuracy & ROC-AUC & $t_{\text{train}}$\\
    \hline
    \hline
    \emph{Adult} & 45,222 & 14 & 111 & 0.81 & 0.85 & 512 & 0.81 & 0.85 & \textbf{115}\\
    \hline
    \emph{ASD} & 248 & 21 & 89 & 0.87 & 0.90 & 198 & 0.87 & 0.90 & \textbf{16}\\
    \hline
    \emph{Cancer} & 569 & 32 & 150 & 0.92 & 0.97 & 168 & 0.92 & 0.94 & \textbf{22}\\
    \hline
    \emph{Heart} & 303 & 13 & 49 & 0.82 & 0.86 & 117 & 0.82 & 0.86 & \textbf{15}\\
    \hline
    \emph{HIV} & 5,840 & 8 & 160 & 0.87 & 0.88 & 449 & 0.87 & 0.88 & \textbf{36}\\
    \hline
    \emph{Titanic} & 2,201 & 3 & 8 & 0.79 & 0.76 & 118 & 0.79 & 0.75 & \textbf{10}
  \end{tabular}
\end{table}

Table~\ref{tab:uci_table} presents the empirical results comparing both implementations.
% The notation in the table follows the definitions in Sec.~\ref{sec:problem_description}.
To guarantee a fair comparison between both implementations we fixed the parameters $r_{\text{max}} = 2$ and $s_{\text{min}} = 0.3$ for both methods, and we set $\mu_{\text{min}} = 0.5$, and $M = 70$ for MCA-miner.
Our multi-core implementations for both MCA-miner and BRL were executed on 6~parallel processes, and only stopped when the Gelman \& Rubin parameter~\cite{brooks1998general} satisfied $\widehat{R} \leq 1.05$.
We ran all the experiments using a single AWS EC2 \verb+c5.18xlarge+ instance with 72~cores.
It is clear from our experiments that our MCA-miner matches the performance of FP-Growth in each case, while significantly reducing the computation time required to mine rules and train BRL models.

\section{Screening Tools for Clinical Psychiatry}
\label{sec:cnp}

The Consortium for Neuropsychiatric Phenomics (CNP)~\cite{poldrack2016a} is a project aimed at understanding shared and distinct neurobiological characteristics among multiple diagnostically distinct patient populations.
Four groups of subjects are included in the study: healthy controls (\emph{HC}, $n=130$), Schizophrenia patients (\emph{SCHZ}, $n=50$), Bipolar Disorder patients (\emph{BD}, $n=49$), and Attention Deficit and Hyperactivity Disorder patients (\emph{ADHD}, $n=43$).
The total number of subjects in the dataset is $n = 272$.
Our goal in analyzing the CNP dataset is to develop \review{interpretable screening tools} to identify the diagnosis of these three psychiatric disorders in patients\review{, as well as finding transdiagnostic tools that identify the commonalities among these disorders}.

\subsection{CNP Self-Reported Instruments Dataset}

Among other data modalities, the CNP study includes responses to $p=578$ individual questions per subject~\cite{poldrack2016a}, belonging to 13 self-report clinical questionnaires with a total of $\sum_{i=1}^p \abs{a_i} = 1350$ categories.
The 13~questionnaires are:
``Adult ADHD Self-Report Screener'' (\emph{ASRS}),
``Barratt Impulsiveness Scale'' (\emph{Barratt}),
``Chapman Perceptual Aberration Scale'' (\emph{ChapPer}),
``Chapman Social Anhedonia Scale'' (\emph{ChapSoc}),
``Chapman Physical Anhedonia Scale'' (\emph{ChapPhy}),
``Dickman Function and Dysfunctional Impulsivity Inventory'' (\emph{Dickman}),
``Eysenck's Impusivity Inventory'' (\emph{Eysenck}),
``Golden \& Meehl's 7 MMPI Items Selected by Taxonomic Method'' (\emph{Golden}),
``Hypomanic Personality Scale'' (\emph{Hypomanic}),
``Hopkins Symptom Check List'' (\emph{Hopkins}),
``Multidimensional Personality Questionnaire -- Control Subscale'' (\emph{MPQ}),
``Temperament and Character Inventory'' (\emph{TCI}), and
``Scale for Traits that Increase Risk for Bipolar II Disorder'' (\emph{BipolarII}).

% \begin{itemize}
% \item Adult ADHD Self-Report Screener (\emph{ASRS}),
% \item Barratt Impulsiveness Scale (\emph{Barratt}),
% \item Chapman Perceptual Aberration Scale (\emph{ChapPer}),
% \item Chapman Physical Anhedonia Scale (\emph{ChapPhy}),
% \item Chapman Social Anhedonia Scale (\emph{ChapSoc}),
% \item Dickman Function and Dysfunctional Impulsivity Inventory (\emph{Dickman}),
% \item Eysenck's Impusivity Inventory (\emph{Eysenck}),
% \item Golden \& Meehl's 7 MMPI Items Selected by Taxonomic Method (\emph{Golden}),
% \item Hopkins Symptom Check List (\emph{Hopkins}),
% \item Hypomanic Personality Scale (\emph{Hypomanic}),
% \item Multidimensional Personality Questionnaire -- Control Subscale (\emph{MPQ}),
% \item Temperament and Character Inventory (\emph{TCI}), and
% \item Scale for Traits that Increase Risk for Bipolar II Disorder (\emph{BipolarII}).
% \end{itemize}
The details about these questionnaires are beyond the scope of this paper, and due to space constraints we abbreviate the individual questions using the name in parenthesis in the list above together with the question number.
% For example, \cnpq{Hopkins}{57} denotes the 57-th question in the ``Hopkins Symptom Check List'' questionnaire.
Depending on the particular clinical questionnaire, each question results in a binary answer (i.e., \true or \false) or a rating integer (e.g., from 1~to 5).
We used each possible answer as a literal attribute, resulting in a range from 2~to 5~categories per attribute.

\subsection{Performance Benchmark}
\label{sec:cnp_perf}

\begin{figure}[t]
  \begin{subfigure}[t]{.32\textwidth}
    \centering
    \includegraphics[width=\linewidth]{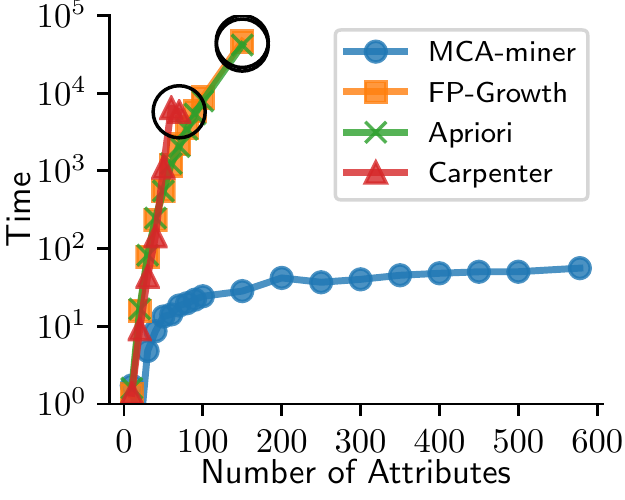}%
    \caption{%
      Rule mining performance comparison.
      % All samples in the plot were obtained by training the same features from the CNP dataset.
      Black circles mark final successful executions.
      % Executions were automatically canceled for wall times longer than 12~hours.
    }%
    \label{fig:run_time_mining}%
  \end{subfigure}%
  \hfill%
  \begin{subfigure}[t]{.32\textwidth}
    \centering
    \includegraphics[width=\linewidth]{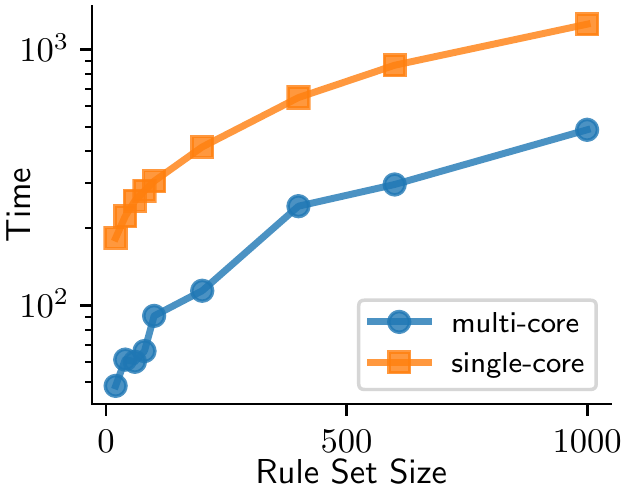}%
    \caption{%
      Convergence run-time of single- and multi-core vs.\ rule set size for 6~MCMC chains.
    }%
    \label{fig:run_time_mcmc}%
  \end{subfigure}%
  \hfill%
  \begin{subfigure}[t]{.32\textwidth}
    \centering
    \includegraphics[width=\linewidth]{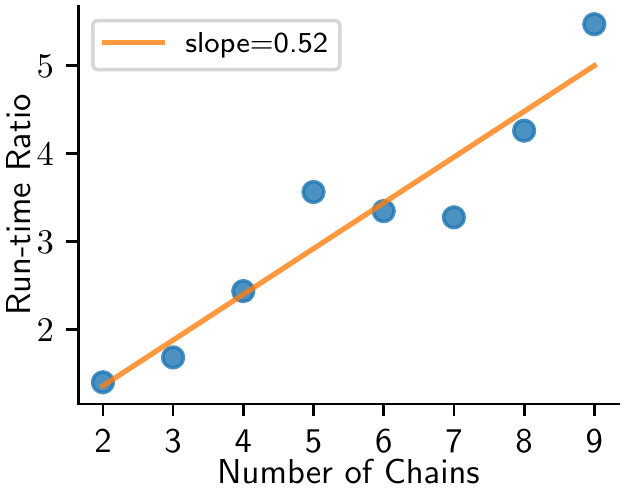}
    \caption{%
      Convergence run-time ratio $\frac{t_{\text{single-core}}}{t_{\text{multi-core}}}$ vs.\ number of multi-core MCMC chains.
      % The number of multi-core used is equal to the number of MCMC chains.
    }%
    \label{fig:mcmc_nchains}%
  \end{subfigure}
  \caption{%
    Wall execution times of our MCA-miner and parallel MCMC implementations.
    All times are an average of 5~runs.
  }%
  \label{fig:run_time}%
\end{figure}

% Rather than prune the number of attributes a priori to reduce the search space for both the rule miner and BRL, we applied our novel MCA-miner to identify the best rules over complete search space of literal combinations.
This is a challenging dataset for most rule learning algorithms since it is \emph{wide}, with more features than samples since $\sum_{i=1}^p \abs{a_i} \gg p \gg n$.
Indeed, just generating all rules with 3~literals from this dataset results in approximately 23~million rules.
Fig.~\ref{fig:run_time_mining} compares the wall execution time of our MCA-miner against three popular associative mining methods: FP-Growth, Apriori, and Carpenter, all using the implementation in the \emph{PyFIM} package~\cite{borgelt2012frequent} and the same set of CNP features.
While the associative mining methods are reasonably efficient on datasets with few features, for datasets with roughly 100~features they result in out-of-memory errors or impractically long executions (longer than 12 hours) even on large-scale compute-optimized AWS EC2 instances.
In comparison, MCA-miner empirically exhibits a grow rate compatible with datasets much larger than CNP.
It is worth noting that while FP-Growth is shown as the fastest associative mining method in~\cite{borgelt2012frequent}, its scaling behavior vs.\ the number of attributes is practically the same as Apriori in our experiments.

In addition to the increased performance due to MCA-miner, we also improved the implementation of the BRL training MCMC algorithm by running parallel Markov chains simultaneously in different CPU cores, as explained in Sec.~\ref{sec:problem_description}.
Fig.~\ref{fig:run_time_mcmc} shows the BRL training time comparison for the same rule set between our multi-core implementation against the original single-core implementation reported in~\cite{letham2015interpretable}.
Also, Fig.~\ref{fig:mcmc_nchains} shows that the multi-core implementation convergence time scales linearly with the number of Markov chains, with $t_{\text{single-core}} \approx \frac{1}{2}\, N_{\text{chains}}\, t_{\text{multi-core}}$.
% While both implementations display a similar grow rate as the rule set size increases, our multi-core implementation is roughly 3~times faster in this experiment.

\subsection{Interpretable Classifiers}

In the interest of building the best possible \review{screening tool for the psychiatric disorders present in the CNP dataset}, we build three different classifiers.
First, we build a binary \review{transdiagnostic} classifier to separate \emph{HC} from the set of \emph{Patients}, defined as the union of \emph{SCHZ}, \emph{BD}, and \emph{ADHD} subjects.
Second, we build a multi-class classifier to separate all four original categorical labels available in the dataset.
Finally, we evaluate the performance of the multi-class classifier \review{as a transdiagnostic tool} by repeating the binary classification task and comparing the results.
All validations were performed using 5-fold cross-validation.
In addition to using Accuracy and ROC-AUC as performance metrics as in Sec.~\ref{sec:uci}, we also report the Cohen's $\kappa$ coefficient~\cite{cohen1960a}, which ranges between -1 (complete misclassification) to 1 (perfect classification), as another indication for the effect size of our classifier since it is compatible with both binary and multi-class classifiers and commonly used in the healthcare literature.
% Cohen's $\kappa$ is commonly used in the field of medical diagnosis and health care for its more robust measure of agreement between predicted labels and actual labels.

\subsubsection{Binary transdiagnostic classifier}

\begin{figure}[t]
  \begin{subfigure}[b]{.67\textwidth}
    \begin{algorithmic}[1]
      \If{\cnpq{ChapSoc}{13} is \true \AlgoAnd \cnpq{Hypomanic}{8} is \true\par%
        \hfill} \emph{Patient} ($P = 0.92$)
      \ElsIf{\cnpq{BipolarII}{1} is \false \AlgoAnd \cnpq{Golden}{1} is \false\par%
        \hfill\AlgoAnd \cnpq{Eysenck}{11} is \false} \emph{HC} ($P = 0.80$)
      \ElsIf{\cnpq{BipolarII}{1} is \false \AlgoAnd \cnpq{Hopkins}{56} is 0\par%
        \hfill} \emph{HC} ($P = 0.50$)
      \ElsIf{\cnpq{BipolarII}{2} is \false \AlgoAnd \cnpq{Hopkins}{39} is 0\par%
        \hfill\AlgoAnd \cnpq{Dickman}{28} is \false} \emph{HC} ($P = 0.50$)
      \Else{} \emph{Patient} ($P = 0.91$)
      \EndIf
    \end{algorithmic}
    \caption{%
      Rule-list for psychiatric transdiagnostic screening.
    }
    \label{fig:cnp_rl_bin_list}
  \end{subfigure}%
  \hfill%
  \begin{subfigure}[b]{.32\textwidth}
    \includegraphics[width=\linewidth]{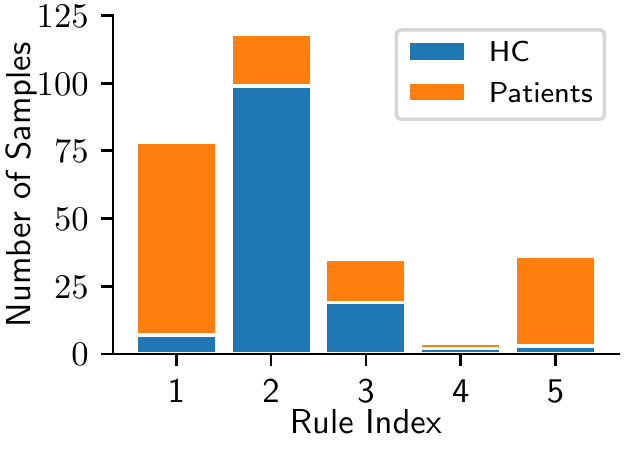}
    \caption{%
      Classification per rule.
    }
    \label{fig:cnp_rl_bin_plot}
  \end{subfigure}
  \caption{%
    Transdiagnostic screening of psychiatric disorders in the CNP dataset.
    Estimated probabilities for each label shown in parenthesis.
  }
  \label{fig:cnp_rl_bin}
\end{figure}

The rule list \review{was generated using all the available samples, namely $130$~\emph{HC} vs.\ $142$~\emph{Patients}, and} is shown in Fig.~\ref{fig:cnp_rl_bin}.
A description of the questions in Fig.~\ref{fig:cnp_rl_bin} is shown in Table~\ref{tab:cnp_questions}.
Note that most subjects are classified with a high probability in the top two rules, which is useful in situations where fast clinical screening is required.
The confusion matrix for this classifier is show in Fig.~\ref{fig:cnp_confusion_2}.

\begin{table}[t]
  \centering
  \caption{%
    Transdiagnostic prediction performance comparison for different models.
  }
  \label{tab:cnp_table}
  \begin{tabular}{l||c|c|c}
    Classifier & Accuracy & ROC-AUC & Cohen's $\kappa$\\
    \hline
    \hline
    MCA-miner + BRL & 0.79 & 0.82 & 0.58\\
    \hline
    Random Forest & 0.75 & 0.85 & 0.51\\
    \hline
    Boosted Trees & 0.79 & 0.87 & 0.59\\
    \hline
    Decision Tree & 0.71 & 0.71 & 0.43
  \end{tabular}
\end{table}

We also benchmark the performance of our method against other commonly used machine learning algorithms compatible with categorical data, using their \emph{Scikit-learn}~\cite{pedregosa2011scikit} implementations and default parameters.
As shown in Table~\ref{tab:cnp_table}, \review{our method has comparable effect size}, if not better, than the state of the art.

\subsubsection{Multi-class classifier}

\begin{figure}[tp]
  \begin{subfigure}[b]{.67\textwidth}
    \centering
    \begin{algorithmic}[1]
      \If{\cnpq{Barratt}{12} is 2 \AlgoAnd \cnpq{Dickman}{29} is \true\par%
        \hfill\AlgoAnd \cnpq{TCI}{231} is \false} \emph{ADHD} ($P = 0.64$)
      \ElsIf{\cnpq{Hypomanic}{1} is \true\par%
        \AlgoAnd \cnpq{Dickman}{22} is \true} \emph{SCHZ} ($P\! =\! 0.72$)
      \ElsIf{\cnpq{BipolarII}{1} is \false \AlgoAnd \cnpq{ChapSoc}{9} is \true\par%
        \hfill} \emph{HC} ($P = 0.70$)
      \Else{} \emph{BD} ($P = 0.44$)
      \EndIf
    \end{algorithmic}%
    \caption{%
      Rule-list for psychiatric screening among 4~classes.
    }
    \label{fig:cnp_rl_list}
   \end{subfigure}%
  \hfill%
  \begin{subfigure}[b]{.32\textwidth}
    \includegraphics[width=\linewidth]{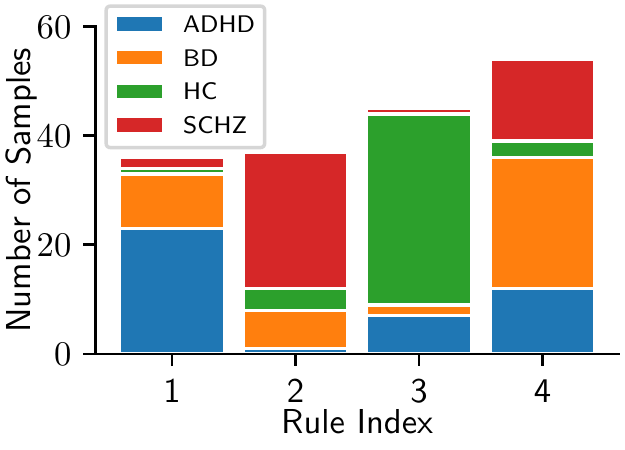}
    \caption{%
      Classification per rule.
    }
    \label{fig:cnp_rl_plot}
  \end{subfigure}
  \caption{%
    Multi-class screening of psychiatric disorders in the CNP dataset.
    Estimated probabilities for each label shown in parenthesis.
  }
  \label{fig:cnp_rl}
\end{figure}

\begin{figure}[t]
  \begin{subfigure}[t]{.32\textwidth}
    \centering
    \includegraphics[width=.8\linewidth]{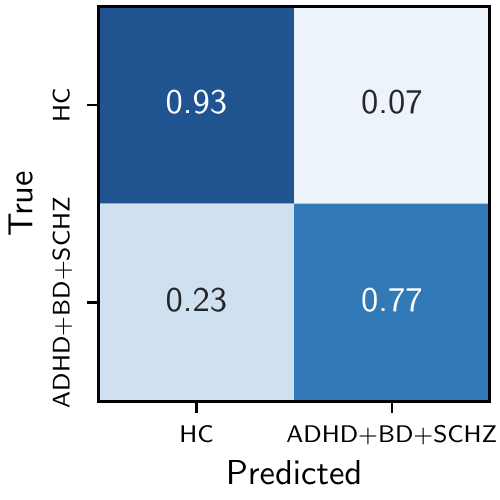}
    \caption{%
      Binary classifier.
    }
    \label{fig:cnp_confusion_2}
  \end{subfigure}%
  \hfill%
  \begin{subfigure}[t]{.32\linewidth}
    \centering
    \includegraphics[width=.8\linewidth]{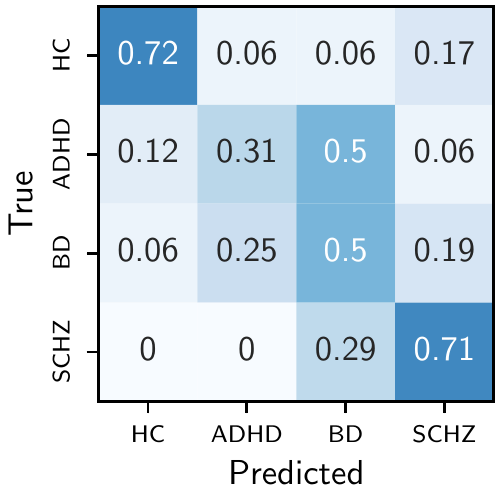}
    \caption{%
      Multi-class classifier.
    }
    \label{fig:cnp_confusion_4}
  \end{subfigure}
  \hfill%
  \begin{subfigure}[t]{.32\linewidth}
    \centering
    \includegraphics[width=.8\linewidth]{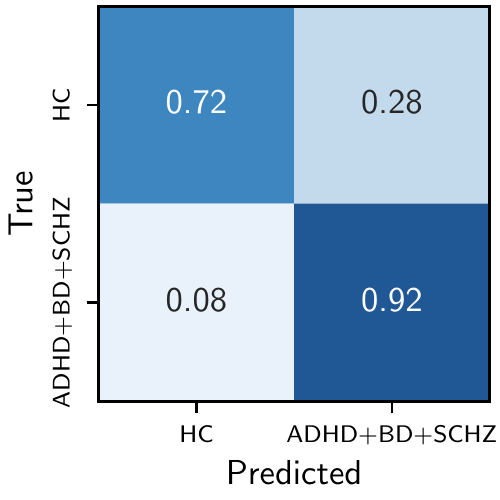}
    \caption{%
      Multi-class with binary emission.
    }
    \label{fig:cnp_confusion_4_2}
  \end{subfigure}
  \caption{%
    Confusion matrices on test cohorts for our classifiers.
  }
  \label{fig:cnp_confusion}
\end{figure}

Fig.~\ref{fig:cnp_rl} shows the rule list obtained using the all 4~labels in the CNP dataset.
\review{We sub-sampled the dataset to balance out each label, resulting in $n=43$ subjects for each of the four classes, with a total of $n=172$ samples.}
% Note that the rules in Fig.~\ref{fig:cnp_rl} emit the maximum likelihood estimate corresponding to the multinomial distribution generated by the same rule in the BRL model, since this is the most useful output for practical clinical use.
Our classifier has an accuracy of $0.57$ and Cohen's $\kappa$ of $0.38$, and Fig.~\ref{fig:cnp_confusion_4} shows the resulting confusion matrix.
The questions present in the rule list are detailed in Table~\ref{tab:cnp_questions}.

% The interpretability and transparency of the rule list in Fig.~\ref{fig:cnp_rl} enables us to obtain further insights regarding the population in the CNP dataset.
% Indeed, similar to the binary classifier, Fig.~\ref{fig:cnp_match_4} shows the mapping of all CNP subjects using the 4-class rule list.
While the accuracy of the rule list as a multi-class classifier is not perfect, it is worth noting how just 7~questions out of a total of~578 are enough to produce a relatively balanced output among the rules.
Also note that, even though each of the 13~questionnaires in the dataset has been thoroughly tested in the literature as clinical instruments to detect and evaluate different traits and behaviors, the 7~questions picked by our rule list do not favor any of the questionnaires in particular.
This is an indication that classifiers are better obtained from different sources of data, and likely improve their performance as other modalities, such as mobile digital inputs, are included in the dataset.

\subsubsection{Binary classification using multi-class rule list}

We replace the \emph{ADHD}, \emph{BD}, and \emph{SCHZ} labels with \emph{Patients} to evaluate the performance of the multi-class classifier as a binary \review{transdiagnostic} classifier.
Using the cross-validated multi-class models, we compute their performance as binary classifiers obtaining an accuracy of $0.77$, ROC-AUC of $0.8$, and Cohen's $\kappa$ of $0.54$.
The confusion matrix is shown in Fig.~\ref{fig:cnp_confusion_4_2}.
These values are on par with those in Table~\ref{tab:cnp_table}, showing that our method does not decrease performance by adding more categorical labels.
\review{Note that while the original binary classifier is highly accurate identifying \emph{HC} subjects, the multi-class classifier with binary emission is better at identifying \emph{Patient} subjects, opening the door to new techniques capable of fusing the best properties of these different rule lists.}

\begin{table}[t]
  \centering
  \caption{%
    CNP dataset questions singled out the rule lists in Figs.~\ref{fig:cnp_rl_bin} and~\ref{fig:cnp_rl}.
    All questions are \true/\false except when noted.
  }
  \label{tab:cnp_questions}
  \begin{tabular}{l|m{283pt}}
    Label & Question \\
    \hline
    \hline
    \cnpq{Barratt}{12} & I am a careful thinker (answer: 1 to 4) \\
    \hline
    \cnpq{BipolarII}{1} & My mood often changes, from happiness to sadness, without my knowing why\\
    \hline
    \cnpq{BipolarII}{2} & I have frequent ups and downs in mood, with and without apparent cause \\
    \hline
    \cnpq{ChapSoc}{9} & I sometimes become deeply attached to people I spend a lot of time with \\
    \hline
    \cnpq{ChapSoc}{13} & My emotional responses seem very different from those of other people \\
    \hline
    \cnpq{Dickman}{22} & I don't like to do things quickly, even when I am doing something that is not very difficult \\
    \hline
    \cnpq{Dickman}{28} & I often get into trouble because I don't think before I act \\
    \hline
    \cnpq{Dickman}{29} & I have more curiosity than most people \\
    \hline
    \cnpq{Eyenseck}{1} & Weakness in parts of your body \\
    \hline
    \cnpq{Golden}{1} & I have not lived the right kind of life \\
    \hline
    \cnpq{Hopkins}{39} & Heart pounding or racing (answer: 0 to 3) \\
    \hline
    \cnpq{Hopkins}{56} & Weakness in parts of your body (answer: 0 to 3) \\
    \hline
    \cnpq{Hypomanic}{1} & I consider myself to be an average kind of person \\
    \hline
    \cnpq{Hypomanic}{8} & There are often times when I am so restless that it is impossible for me to sit still \\
    \hline
    \cnpq{TCI}{231}  & I usually stay away from social situations where I would have to meet strangers, even if I am assured that they will be friendly
  \end{tabular}
\end{table}

\section{DISCUSSION}
\label{sec:discuss}

% We propose a novel method to analyze categorical datasets with a large number of attributes, a property that is prevalent in the clinical psychiatry community.
We formulated a novel MCA-based rule mining method, with excellent scaling properties against the number of categorical attributes, and presented a new implementation of the BRL algorithm using multi-core parallelization.
We also studied the CNP dataset for psychiatric disorders using our new method, resulting in rule-based interpretable classifiers capable of screening patients from self-reported questionnaire data.
Our results not only show the viability of building interpretable models for state-of-the-art clinical psychiatry datasets, but also highlight the scalability of these models to larger datasets to understand the interactions and differences between these disorders.
\review{We are actively exploring avenues for improving recruitment and reducing screening rejections in clinical trials.}

% \input{sections/reviews}
% \input{sections/supplementary}
% \input{sections/acknowledgements}

%%%%%
%% References
%%%%%
\bibliographystyle{splncs04}
\bibliography{references}

\end{document}